\documentclass{article}
\usepackage{subfigure}

\usepackage[ForAI,final]{neurips_2025} 

\usepackage[utf8]{inputenc}
\usepackage[T1]{fontenc}
\usepackage{amsmath, amssymb, amsfonts}
\usepackage{booktabs}
\usepackage{nicefrac}
\usepackage{microtype}
\usepackage{adjustbox}
\usepackage{xspace}
\usepackage[dvipsnames]{xcolor}
\usepackage{graphicx}
\usepackage{subcaption}
\usepackage{caption}
\usepackage{wrapfig}
\usepackage{enumitem}
\usepackage{array}

\usepackage{hyperref}
\hypersetup{colorlinks=true, allcolors=blue}

\newcommand{\ie}{\emph{i.e.}}
\newcommand{\eg}{\emph{e.g.}}

\title{Towards Integrating Uncertainty \\ for Domain-Agnostic Segmentation}

\author{%
  Jesse Brouwers 
  \qquad Xiaoyan Xing
  \qquad Alexander Timans \\
  UvA-Bosch Delta Lab, University of Amsterdam
}

\begin{document}

\maketitle

\begin{abstract}
    Foundation models for segmentation such as the Segment Anything Model (SAM) family exhibit strong zero-shot performance, but remain vulnerable in shifted or limited-knowledge domains. This work investigates whether uncertainty quantification can mitigate such challenges and enhance model generalisability in a domain-agnostic manner. To this end, we (1) curate \texttt{UncertSAM}, a benchmark comprising eight datasets designed to stress-test SAM under challenging segmentation conditions including shadows, transparency, and camouflage; (2) evaluate a suite of lightweight, \emph{post-hoc} uncertainty estimation methods; and (3) assess a preliminary uncertainty-guided prediction refinement step. Among evaluated approaches, a last-layer Laplace approximation yields uncertainty estimates that correlate well with segmentation errors, indicating a meaningful signal. While refinement benefits are preliminary, our findings underscore the potential of incorporating uncertainty into segmentation models to support robust, domain-agnostic performance. Our benchmark and code are made available at \url{https://github.com/JesseBrouw/UncertSAM}.
\end{abstract}

\section{Introduction}
\label{sec:intro}

\begin{wrapfigure}{r}{0.55\textwidth}
\vspace{-10pt}
\centering

\begin{minipage}[b]{0.48\linewidth}
  \includegraphics[width=\linewidth]{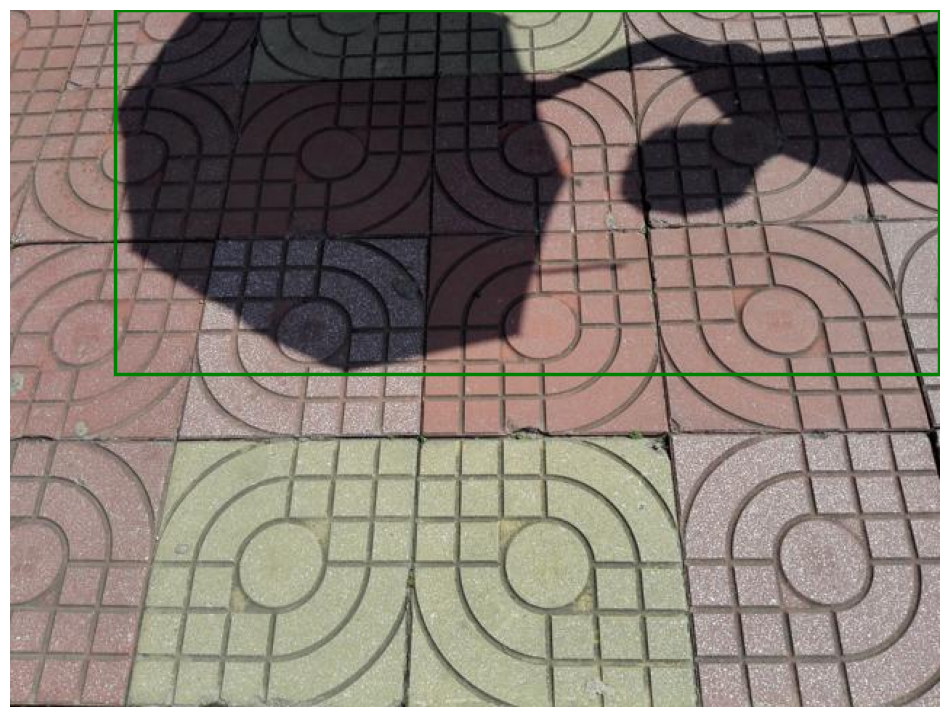}
  \captionof{subfigure}{Target \& box prompt}
\end{minipage}\hfill
\begin{minipage}[b]{0.48\linewidth}
  \includegraphics[width=\linewidth]{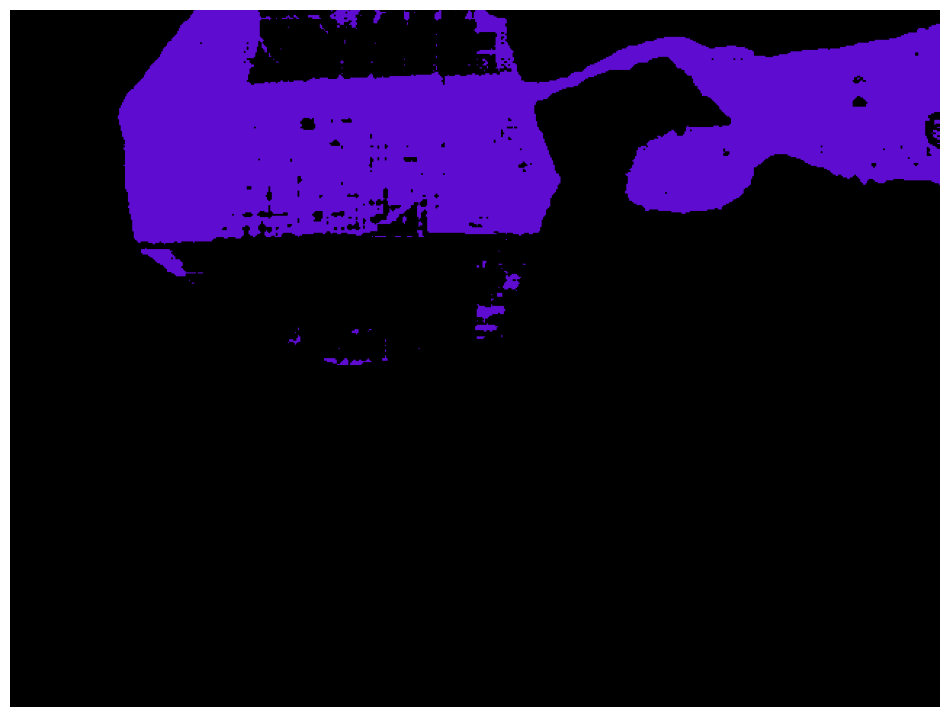}
  \captionof{subfigure}{Predicted mask}
\end{minipage}

\vspace{2mm}

\begin{minipage}[b]{0.48\linewidth}
  \includegraphics[width=\linewidth]{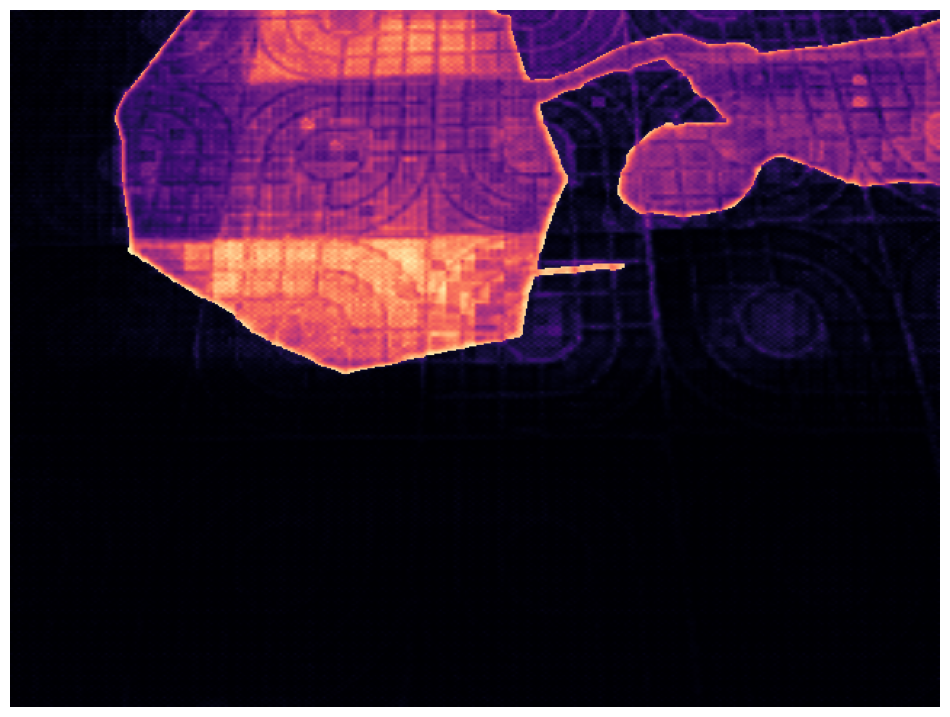}
  \captionof{subfigure}{Prob. error map}
\end{minipage}\hfill
\begin{minipage}[b]{0.48\linewidth}
  \includegraphics[width=\linewidth]{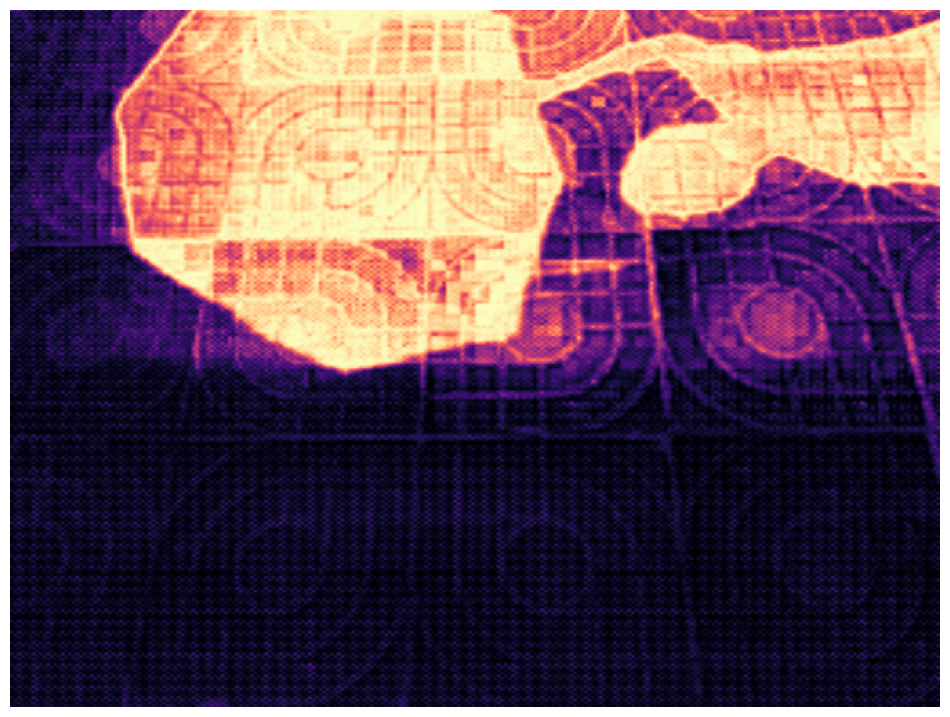}
  \captionof{subfigure}{Uncertainty map}
\end{minipage}

\caption{\textbf{An example of SAM's failure in shadow detection.} 
The predicted mask misses shadow regions despite an accurate bounding box prompt. 
In contrast, a Laplace-based uncertainty map correctly recovers the full shadow (from the ISTD dataset \citep{wang2018stacked}).}

\label{fig:istd_ignorance}
\vspace{-20pt}
\end{wrapfigure}

As in other domains, large-scale foundation models have transformed vision-based tasks and enabled effective generalisation to novel tasks through zero- or few-shot prompting \citep{bommasani2021opportunities}. For segmentation tasks, the Segment Anything Model family (SAM) stands out through its high-quality segmentation masks leveraging minimal user input, such as point clicks or bounding boxes \citep{kirillov2023segment}. Trained on the SA-1B dataset of over \emph{one billion} masks, the second SAM iteration \citep{ravi2024sam} demonstrates exceptional generalisation, yet struggles with fine-grained structures, precise object boundaries, and sensitivity to common real-world degradations such as motion blur, noise, shadows, and transparency \citep{kirillov2023segment, ji2024segment, wang2024empirical, chen2024robustsam}. 
Uncertainty quantification (UQ) offers a potential avenue to enhance SAM's robustness to such cases. By supplementing predictions with an added measure of uncertainty or confidence, UQ can help raise the trustworthiness of segmentation outputs and notify the model when it risks being wrong \citep{gawlikowski2023survey}. However, current efforts to incorporate uncertainty into SAM are limited in scope. Most existing studies focus on narrow task settings or estimate uncertainty purely via heuristics such as boundary cues, rather than eliciting it more fundamentally from the model \citep{zhang2023segment, zhou2025medsam, liu2024uncertainty, kaiser2025uncertainsam, xie2024pa}. Thus, it remains somewhat unclear whether meaningful spatial uncertainty estimates can be obtained, and whether these can serve as a signal to refine predictions in a general, domain-agnostic fashion (see \autoref{fig:istd_ignorance} for a visual example). 

In this work, we address these questions by comparing four approaches to quantify pixel-level segmentation uncertainty via controlled perturbations to the input, prompts, model parameters, or an additional variance network (\autoref{fig:uq-methods}). Our approaches are driven by the desire for a lightweight, \emph{post-hoc} uncertainty integration amenable to work with a pretrained and \emph{frozen} SAM encoder\footnote{We refer to SAM more broadly, but in practice work with SAM-2 \citep{ravi2024sam}.}. While all methods yield reasonable uncertainties, we find that a last-layer Laplace approximation most strongly correlates with segmentation errors, highlighting its potential to guide prediction refinement.

We take a first step towards a refinement strategy via a dense embedding that fuses uncertainty maps into SAM's encoder representations (\autoref{app:dense-embedding-fusion}), but minor gains over control baselines suggest it falls short of fully exploiting the uncertainty signal. We posit this originates from limitations of our purely \emph{post-hoc} approach, and stipulate a deeper integration of uncertainty into model architecture to improve performance benefits. In summary, we contribute:

\begin{itemize}[leftmargin=12pt]
\item \texttt{UncertSAM}, a curated multi-domain benchmark featuring challenging segmentation cases and enabling domain-agnostic evaluation of methods; 
\item a systematic comparison of four pixel-level uncertainty methods for SAM, showing strong alignment between uncertainty and segmentation errors;
\item a simple prediction refinement strategy leveraging uncertainty estimates, in part achieving small gains while necessitating no domain-specific fine-tuning.
\end{itemize}


\section{Methodology}
\label{sec:method}

We next detail our three-step approach on benchmarking, uncertainty estimation and refinement.

\paragraph{Benchmark curation.} We establish the \texttt{UncertSAM} benchmark by collecting and standardising eight datasets spanning a range of challenging visual conditions and environments for segmentation. These include fine-grained salient objects (BIG~\citep{cheng2020cascadepsp}, COIFT~\citep{liew2021deep}), camouflaged objects (COD~\citep{fan2022concealed}), medical CT scans (MSD Spleen~\citep{antonelli2022medical}), shadows (ISTD~\citep{wang2018stacked}, SBU~\citep{vicente2016large}), lighting artifacts (Flare~\citep{dai2022flare7k}), and transparent objects (Trans~\citep{xie2021segmenting}). This results in a collection of over 23,000 images and 44,000 annotated masks across different domains and edge cases, which can be used to evaluate baseline segmentation performance, UQ methods, and uncertainty-guided refinement. To ensure domain-agnostic analysis, any uncertainty fitting (for the Laplace) or training (for the variance network) is done on a representative subset of SAM's original training set (SA-1B~\citep{kirillov2023segment}). Further dataset details are provided in \autoref{app:uncertsam}.

\begin{figure}[!t]
    \centering
    \includegraphics[width=\linewidth]{images/UQ_method_diagram_cropped.png}
    \caption{
    \textbf{Overview of our four uncertainty strategies.} \emph{From left to right:} We quantify pixel-level uncertainty by targeting the input image (via test-time augmentations, TTA), prompts, model parameters (via the Laplace approximation, LA), and an additional variance prediction head. The first three methods leverage stochasticity to generate ensembles, whereas the latter is deterministic.
    }
    \vspace{-4mm}
    \label{fig:uq-methods}
\end{figure}

\paragraph{Uncertainty quantification for SAM.} We consider four UQ strategies which target different components of the model design, thus providing complementary views on arising uncertainty (see \autoref{fig:uq-methods}). Since we refrain from modifying the parameter-heavy SAM image encoder, our approaches are generally \emph{post-hoc} and target the lightweight SAM prompt and decoder modules. We refer to \cite{ravi2024sam} for an architectural overview of SAM and its three key components. Consider the input combination of an image ${X \in \mathbb{R}^{H \times W \times C}}$ and prompt configuration $q$, such as user-provided bounding box coordinates $q \in \mathbb{R}^4$ around the target object to segment. We generically define $f_{\theta}$ as the parametrised SAM model with weights $\theta$, and subsequently $f_{\theta}(X, q) \in \mathbb{R}^{H \times W}$ as the model's pixel-wise output logits for the input tuple $(X, q)$. As SAM targets binary foreground/background segmentation, a sigmoid function $\sigma$ can be applied pixel-wise to obtain a final probability map $P \in [0,1]^{H \times W}$. With this notation in hand, we employ the following four uncertainty strategies:

\noindent \emph{(i) Test-Time Augmentations (TTA).} We perturb the input image by sampling and applying a stochastic augmentation $T_n \sim \mathcal{T}$, resulting in the probability map $P_n = \sigma(f_\theta(T_n(X), q)).$ We consider augmentations used during SAM's original training (\eg~flips, resizes, jitter; see \autoref{tab:augmentation_settings}) as well as additional hue shifts. Sampling $N$ times yields a set of maps $\{P_1, \dots, P_N\}$, which can then be used to obtain a pixel-wise mean prediction map $\bar{P} = \frac{1}{N} \sum_{n=1}^N P_n$ and uncertainty map $U$. Among different options to measure uncertainty we consider the pixel-wise \emph{predictive entropy} $\mathbb{H}[\cdot]$, given for the binary case as $ U = \mathbb{H}[\bar{P}] = -\,\bar P\log\bar P - (1-\bar P)\log(1-\bar P)$. We refer to this as our \emph{predictive} spatial uncertainty map, rather than claiming distinct origins \citep{hullermeier2021aleatoric}.

\noindent\emph{(ii) Prompt Perturbations.} Instead of augmenting the input image, we may also perturb the input prompt by sampling bounding box coordinate perturbations $R_n \sim \mathcal{R}$ and obtaining the probability map $ P_n = \sigma(f_\theta(X, R_n(q)))$. We leverage the existing prompt perturbation schedule used during SAM's training \citep{kirillov2023segment, ravi2024sam}, and generate an ensemble mean $\bar{P}$ and uncertainty map $U$ as above.

\noindent\emph{(iii) Last-layer Laplace Approximation (LA).} In order to generate an ensemble from model parameters, we consider a scalable Bayesian treatment via the Laplace approximation over the final linear decoder layer \citep{mackay1992practical, daxberger2021laplace}. A Gaussian posterior approximation is centred over the layer's pretrained model weights---interpretable as the \emph{maximum a posteriori} estimates $\hat{\theta}_{\text{MAP}}$---with variance dictated by the local curvature of the (diagonal) Hessian $\hat{H}$, that is $p(\theta \mid \mathcal{D}_{\text{fit}}) \approx \mathcal{N}(\theta \mid \hat{\theta}_{\text{MAP}}, \hat{H}^{-1}).$ This offers a relatively crude, but scalable and \emph{post-hoc} Bayesian model treatment, and model weights can now simply be sampled as $\theta_n \sim p(\theta \mid \mathcal{D}_{\text{fit}})$ to produce the probability map $P_n = \sigma(f_{\theta_n}(X, q))$ and obtain $\bar{P}$ and $U$ as before.

\noindent\emph{(iv) Learnable variance network.} Finally, a more distinct approach inspired by \cite{kendall2017uncertainties} sees training an auxiliary uncertainty prediction head on top of SAM's decoder features. The head learns to predict a pixel-wise log variance term, and is interpretable as a Gaussian `spread' given the trained likelihood objective (see \cite{kendall2017uncertainties} and \autoref{app:variance_network_architecture} for more details). Both the decoder and mean prediction are kept frozen, and thus only a notion of uncertainty is learned in a strictly \emph{post-hoc} way. In contrast to above, the returned uncertainty map $U$ is not based on ensembling but comes from a deterministic variance prediction given the inputs $(X, q)$. 

\paragraph{Uncertainty-guided prediction refinement.} How can the obtained map $U$ subsequently be used to improve predictions and help mitigate failures such as \autoref{fig:istd_ignorance}? In this work, we consider a simple approach dubbed \emph{Dense Embedding Fusion}, which encodes both prediction and uncertainty into a dense embedding and applies a \(1 \times 1\) convolution to produce an uncertainty-aware fused feature map. Repeating a second forward pass through SAM, we expect this map to supplement the decoder's internal spatial features with uncertainty information to aid correct initial mistakes. See \autoref{app:dense-embedding-fusion} for a schematic and details. We stress that this is a \emph{preliminary first step} towards more elaborate strategies.

\begin{table*}[!t]
\centering
\setlength{\tabcolsep}{10pt}
\begin{adjustbox}{max width=\linewidth}
\begin{tabular}{lcccccccc}
\toprule
  & \textbf{BIG} & \textbf{COIFT} & \textbf{COD} & \textbf{MSD} & \textbf{ISTD} & \textbf{SBU} & \textbf{Flare} & \textbf{Trans} \\
\midrule
    \multicolumn{9}{l}{\textbf{Mean IoU} ($\uparrow$)} \\
    \textit{Ground truth mask} & \textit{93.96} & \textit{95.65} & \textit{89.20} & \textit{91.35} & \textit{92.52} & \textit{87.09} & \textit{82.15 }& \textit{94.92} \\
\midrule 
    SAM (No Refinement) & \textbf{89.95} & \textbf{94.77} & \textbf{82.82} & \underline{87.86} & 65.71 & 66.11 & 42.82 & 84.87 \\
    Dense Fusion w/ SAM (Ones Map)  & \underline{89.82} & 94.73 & \underline{82.72} & \textbf{87.91} & 66.53 & 66.27 & 43.12 & 85.05 \\
    Dense Fusion w/ LA (Ones Map) & 89.34 & \underline{94.74} & 82.54 & 87.25 & \textbf{67.89} & \textbf{67.37} & \textbf{46.61} & \textbf{86.54} \\
    Dense Fusion w/ LA & 88.58 & 94.73 & 82.41 & 87.49 & \underline{67.70} & \underline{67.26} & \underline{46.19} & \underline{86.49} \\
\midrule
    \multicolumn{9}{l}{\textbf{Mean Boundary IoU} ($\uparrow$)} \\
    \textit{Ground truth mask}         & \textit{86.50 }& \textit{91.25} & \textit{83.20} & \textit{87.76 }& \textit{76.92} & \textit{80.34} & \textit{72.36} & \textit{85.15} \\
\midrule
    SAM (No Refinement) & \textbf{83.81} & \textbf{89.74} & \textbf{75.01} & \underline{82.69} & 47.71 & 57.49 & 35.47 & 69.82 \\
    Dense Fusion w/ SAM (Ones Map)   & 83.34 & \textbf{89.74} & \underline{74.86} & \textbf{82.71} & 48.17 & 57.62 & 35.32 & 69.75 \\
    Dense Fusion w/ LA (Ones Map)      & \underline{83.40} & 89.58 & 74.47 & 81.84 & \textbf{48.92} & \textbf{58.04} & \textbf{36.71} & \textbf{72.31} \\
    Dense Fusion w/ LA & 82.40 & 89.60 & 74.46 & 82.21 & \textbf{48.92} & \underline{57.98} & \underline{36.48} & \underline{72.28} \\
\bottomrule
\end{tabular}
\end{adjustbox}
\caption{
\textbf{Comparison of uncertainty-guided prediction refinement} across the \texttt{UncertSAM} benchmark. \textit{Ground truth mask} represents an empirical upper bound using the ground truth mask for refinement. \textit{SAM (No Refinement)} refers to the baseline SAM prediction without refinement step. \textit{Dense Fusion w/ SAM} applies dense embedding fusion with baseline SAM prediction maps, whereas \textit{Dense Fusion w/ LA} uses Laplace-based prediction and uncertainty maps. Variants marked \textit{(Ones Map)} explicitly fuse a constant map of ones instead of uncertainty maps. \textbf{Best values}, \underline{second best}.
}
\label{tab:prompt_refinement}
\vspace{-6mm}
\end{table*}

\vspace{-2mm}
\section{Experimental Results}
\label{sec:exp}

\begin{wrapfigure}{r}{0.53\textwidth}
\centering
\includegraphics[width=\linewidth]{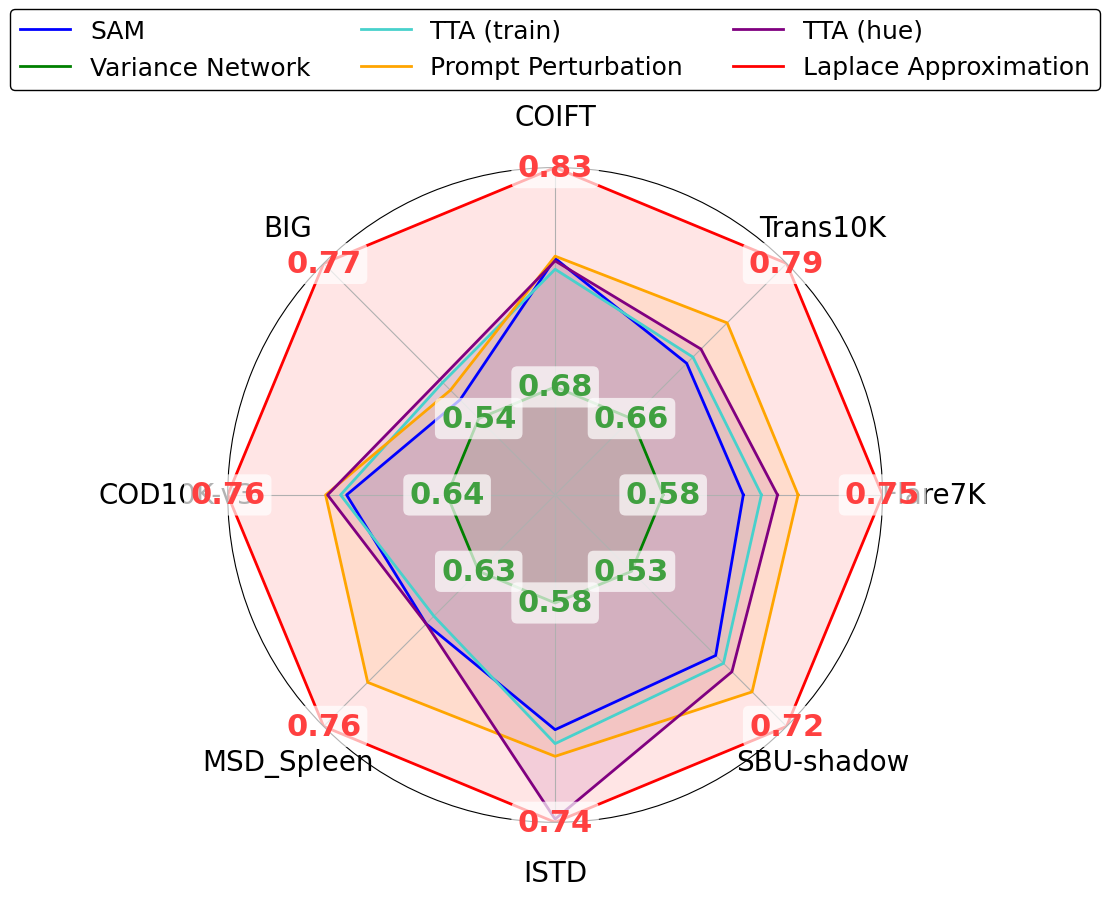}
\caption{
\textbf{Pearson correlation coefficient} $\rho$ between error maps and uncertainty maps for each method, averaged across samples per dataset. Higher positive values indicate better correlation, with the LA giving strongest error alignment.
}
\label{fig:correlation_with_error}
\vspace{-15pt}
\end{wrapfigure}

We assess UQ methods by their correlation with prediction error in \autoref{fig:correlation_with_error}, and then test their utility for downstream refinement in \autoref{tab:prompt_refinement}. All experiments make use of bounding box prompts only.

\paragraph{Uncertainty alignment with error.} We measure alignment via the \emph{Pearson correlation coefficient} $\rho(U, E)$, where $E = |P - M|$ denotes the probabilistic model error, \ie~the gap between probability map $P$ and ground-truth foreground/background segmentation mask $M \in [0,1]^{H \times W}$. A larger positive value indicates stronger linear correlation between $U$ and $E$ \citep{benesty2009pearson}. Our results in \autoref{fig:correlation_with_error}, averaged across pixels and samples for each dataset, find that correlation is generally high across methods ($\rho > 0.5$), indicating good correspondence. The variance network records lowest values, whereas the Laplace Approximation gives strongest alignment. Thus we focus particularly on this approach for subsequent refinement. Similar analysis via the \emph{Brier score} metric \citep{gneiting2007strictly} indicated comparable probabilistic accuracy across methods (\autoref{fig:brier_scores}), and more visuals are given in \autoref{app:more-results}.

\vspace{-2mm}
\paragraph{Utility for prediction refinement.} We measure predictive performance via two standard segmentation metrics, \emph{mean intersection-over-union} (mIoU) and its boundary-only version (mBIoU) which focuses on pixels exclusively on the mask contours, following \cite{ke2023segment}. We benchmark our uncertainty-guided refinement step (\emph{Dense Fusion w/ LA}) against multiple controls to isolate sources of potential gains, detailed further in \autoref{app:exp-design}. We also evaluate SAM directly, and report an additional empirical \emph{performance upper bound} leveraging the ground truth mask for refinement. We observe mixed results across the \texttt{UncertSAM} benchmark in \autoref{tab:prompt_refinement}. We find that leveraging the LA's ensemble prediction improves over SAM's baseline predictions, in particular for more challenging domain-shifted datasets (ISTD, SBU, Flare and Trans). However, our dense fusion approach does not improve meaningfully over controls (Ones Map), suggesting our explicit uncertainty handling in its current preliminary form does not contribute consistently to refine predictions. 

\vspace{-2mm}
\paragraph{Conclusion.} We observe the recovery of meaningful uncertainty estimates that can benefit predictions, but a simple fusion approach fails to fully leverage this uncertainty signal. In part, this may be due to our \emph{post-hoc} driven strategies which freeze SAM's image encoder, containing most of its expressivity. A deeper fusion of uncertainty into the model architecture, or even the learning process itself, should yield better leverage for subsequent refinement. Nonetheless, we believe such a more holistic perspective on using uncertainty to guide predictions, as opposed to pure per-domain fine-tuning, can offer a more principled path to robust and reliable domain-agnostic segmentation. 

\bibliographystyle{plainnat}
\bibliography{main}

\clearpage
\appendix
\begin{center}
    {\Large\bfseries Towards Integrating Uncertainty for Domain-Agnostic Segmentation}\\[0.5em]
    {\Large\bfseries --- Supplementary Material ---}
\end{center}

\section{Further Experimental Design \& Discussion}
\label{app:exp-design}

We measure predictive performance via two standard segmentation metrics, \emph{mean intersection-over-union} (mIoU) and its boundary-only version (mBIoU) which focuses on pixels exclusively on the mask contours. Following \cite{ke2023segment} the boundary distance is set dynamically based on image size. All experiments are conducted on NVIDIA H100 GPUs (80GB VRAM) with CUDA 12.6.0.

\paragraph{Control baselines.} In order to thoroughly benchmark our uncertainty-guided refinement step (Dense Fusion w/ LA), we compare against two controls to isolate sources of potential gains. To verify if explicit fusion of the uncertainty map is useful, we compare to fusion with an uninformative constant map instead (Dense Fusion w/ LA, Ones Map). To verify if benefits are gained from using the LA's ensemble predictions, we compare to a fusion using constant maps \emph{and} SAM's baseline prediction (Dense Fusion w/ SAM, Ones Map). We also evaluate SAM directly, and report an additional empirical \emph{performance upper bound} leveraging the ground truth mask for refinement. 

\paragraph{Results analysis.} Our results across the \texttt{UncertSAM} benchmark in \autoref{tab:prompt_refinement} draw mixed conclusions. On one hand, fusing the uncertainty map \emph{does not} improve over its constant control (Ones Map), suggesting that the explicit uncertainty handling using our approach contributes little to refine predictions. On the other hand, leveraging the LA's ensemble prediction \emph{does} improve over SAM's baseline predictions, in particular for more challenging domain-shifted datasets (ISTD, SBU, Flare and Trans)\footnote{In part, this may also stem from the introduced fusion layer's ability to re-weigh latent features in a way that benefits generalisation.}. However, the gap to the ground-truth upper bound across these datasets remains fairly large across the board. In contrast, highly in-domain (but fine-grained) datasets such as BIG and COIFT see strong performance even from baseline SAM. Overall, we observe that meaningful uncertainty estimates are recovered and can certainly benefit predictions, but our simple fusion approach fails to fully leverage this uncertainty signal for explicit prediction refinement.

\paragraph{Discussion.} Uncertainty estimates that correlate well with model error are meaningful, and should help the model refine its predictions especially in high-error regions (\autoref{fig:istd_ignorance}). Yet, our control experiments suggest that prediction benefits originate primarily from improved ensemble predictions, rather than explicit uncertainty information passed on through dense embedding fusion. Despite modest gains, our prediction refinement step is hindered by two major limitations. 

Firstly, our \emph{post-hoc} driven strategies operate by freezing SAM's image encoder, which contains most of its expressivity. Thus, the model may lack capacity to adapt necessary internal representations in response to uncertainty signals. A deeper fusion of uncertainty into the model architecture, or even the learning process itself, should yield better leverage for subsequent refinement. Secondly, our design prioritised domain-agnostic generalisation, and as such any uncertainty fitting or training is done on in-domain training data (from SA-1B \citep{kirillov2023segment}) containing predominantly high-confidence examples. This will have constrained the refinement module’s exposure to different lower-confidence uncertainty patterns arising in domain-shifted settings. A more balanced regime, or cross-domain fine-tuning exposure could help the model develop a richer representation of uncertainty, benefitting downstream generalisation.

Moving forward, these directions can be explored to shift from a strictly \emph{post-hoc} perspective toward uncertainty-aware model integration and learning.

\section{UncertSAM benchmark}
\label{app:uncertsam}

\begin{table}[t]
\centering
\caption{Datasets overview of the \texttt{UncertSAM} benchmark, including licensing information and pre-processing configurations. The columns indicate pre-processing steps on connected component analysis (CCA), colour-coding (CC), and specific medical CT scan pre-processing.}
\vspace{3mm}
\begin{adjustbox}{max width=\linewidth}
\begin{tabular}{@{}l l l c c c@{}}
\toprule
\textbf{Dataset} & \textbf{Reference} & \textbf{License} & \textbf{CCA} & \textbf{CC} & \textbf{CT} \\
\midrule
\texttt{BIG}       & \cite{cheng2020cascadepsp} & \texttt{Research Only} & \checkmark &  &  \\
\texttt{COIFT}     & \cite{liew2021deep} & \texttt{Attribution-NonCommercial 4.0} &  &  &  \\
\texttt{COD10K-v3} & \cite{fan2022concealed} & \texttt{Research Only} & \checkmark & \checkmark &  \\
\texttt{MSD Spleen} & \cite{antonelli2022medical} & \texttt{Attribution-ShareAlike 4.0} &  &  & \checkmark \\
\texttt{ISTD}      & \cite{wang2018stacked} & \texttt{Research Only} & \checkmark &  &  \\
\texttt{SBU}       & \cite{vicente2016large} & Unknown & \checkmark &  &  \\
\texttt{Flare7K}   & \cite{dai2022flare7k} & \href{https://github.com/ykdai/Flare7K/blob/main/LICENSE}{S-Lab License 1.0} & \checkmark & \checkmark &  \\
\texttt{Trans10K}  & \cite{xie2021segmenting} & \texttt{Research Only} & \checkmark &  &  \\
\midrule 
\texttt{SA-1B Subset} & \cite{kirillov2023segment} & \href{https://ai.facebook.com/datasets/segment-anything}{SA-1B V1.0} & &  &  \\
\bottomrule
\end{tabular}
\end{adjustbox}
\label{tab:uncertsam_overview}
\vspace{-3mm}
\end{table}

\autoref{tab:uncertsam_overview} contains an overview of the datasets used in the \texttt{UncertSAM} benchmark. Dataset licenses, where available, are included. Most licenses restrict usage to research purposes only. The SA-1B dataset \citep{kirillov2023segment} has a more restrictive license, and therefore it is not included in our publicly available dataset. However, original filenames from the randomly sampled subset used in this study are retained by the author. For reproducibility, this subset is available upon request. The preprocessing steps indicated in the table columns are described below:

\begin{itemize}[leftmargin=12pt]
    \item \textbf{Connected Component Analysis (CCA).} For datasets containing images with multiple disconnected surfaces potentially representing valid masks according to SAM’s broad entity-part strategy, we apply CCA to separate masks. We use the \texttt{OpenCV} library to perform the following steps:
    \begin{enumerate}
        \item Apply morphological closing twice to the initial mask using a $3\times3$ kernel to reduce the likelihood of generating semantically irrelevant masks when small parts are disconnected but belong to a larger coherent target.
        \item Extract connected components from the closed mask.
        \item Perform a binary AND operation between the resulting mask and the initial mask to remove regions introduced by morphological closing.
        \item Retain connected components larger than 1,000 pixels to eliminate small artifacts.
    \end{enumerate}
    \item \textbf{Colour Coded (CC).} When masks are colour coded, we extract unique RGB values and split the mask accordingly into multiple targets.
    \item \textbf{CT scan processing.} The CT images in the \texttt{MSD Spleen} dataset are 3D CT volumes in \texttt{nii.gz} format, sliced along the axial plane to produce 2D images. Only slices containing foreground labels are retained. Normalisation follows the method described in \cite{du2025segvol}:
    \begin{enumerate}
        \item Filter the volume to keep only foreground voxels.
        \item Apply Z-score normalisation: $\frac{x - \mu}{\sigma}$.
        \item Clip voxel intensities to the [0.05, 99.95] percentile range.
    \end{enumerate}
\end{itemize}

\section{Variance Network Architecture}
\label{app:variance_network_architecture}

\begin{figure}[!t]
    \centering
    \includegraphics[width=\linewidth]{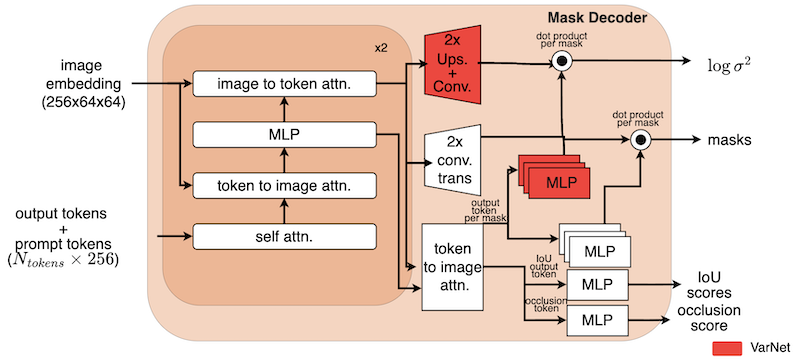}
    \caption{
        Modified SAM mask decoder architecture with an auxiliary variance network.
    }
    \label{fig:variance_network_architecture}
\end{figure}

The auxiliary variance network deterministically predicts log variance by adding two components to the SAM mask decoder: a variance prediction head, identical to the existing mask prediction heads, and a lightweight CNN that upsamples spatial embeddings from the image-to-token attention. To reduce checkerboard artifacts in the variance outputs, we replace transpose convolutions in the CNN with bilinear upsampling followed by 2D convolution. \autoref{fig:variance_network_architecture} shows the modified mask decoder.

\begin{figure*}[!ht]
    \centering
    \includegraphics[width=\linewidth]{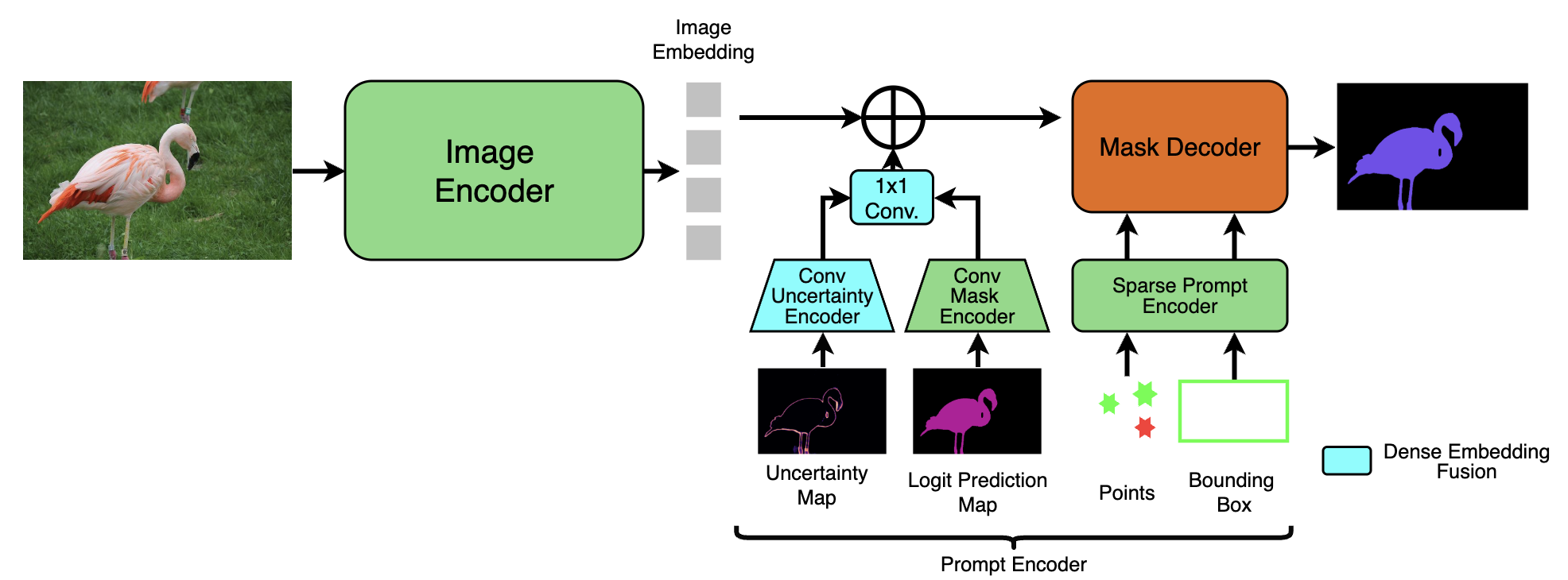}
    \caption{Architecture of the Dense Embedding Fusion Network. The prompt encoder is extended with a CNN-based uncertainty encoder, operating in parallel to the mask prompt encoder used in SAM models. The resulting dense spatial embeddings are concatenated along the channel dimension and fused via a $1 \times 1$ convolution layer.}
    \label{fig:refinement_step}
\end{figure*}

\section{Dense Embedding Fusion Architecture}
\label{app:dense-embedding-fusion}
We extend SAM's prompt encoder with a parallel uncertainty embedding module that processes a spatial uncertainty map to produce dense features. These are concatenated with the mask prompt embeddings and fused via a $1 \times 1$ convolution to retain the original channel dimensionality. This fused representation replaces the original mask embedding, allowing uncertainty to be integrated directly into the internal spatial features used by the decoder. An overview of the architecture is shown in \autoref{fig:refinement_step}.

\section{Training and Hyperparameter Configurations}
\label{app:training_and_fitting}

All experiments are conducted on NVIDIA H100 GPUs (80GB VRAM) with CUDA 12.6.0. All post-hoc fitting and training uses the SA-1B subset to enable a domain-agnostic analysis.

\paragraph{Multi-step training schedule.} We follow an 8-step schedule similar to SAM~\citep{ravi2024sam}. The first prompt is sampled as a bounding box or a single foreground point with equal probability. Each subsequent step adds one point, sampled uniformly from foreground or background. Unlike the setup of SAM, which places new points in error regions to simulate interactive correction, our uniform sampling targets broad prompt diversity to better characterise uncertainty rather than maximise iterative correction quality.

\paragraph{Fitting the Laplace Approximation.} We use the SAM model with frozen weights and fit a diagonal Hessian approximation over the final layer. We fit the LA for one epoch with a batch size of 1. Due to memory constraints, we sample one of the eight prompts per optimisation step to maintain diversity while keeping memory usage low. The loss is computed on predictions and masks downsampled to $128\times128$ because of memory constraints.

\paragraph{Variance network training.} This approach adds a variance head on top of the SAM backbone, which remains frozen. The model is trained using the heteroscedastic loss proposed by~\cite{kendall2017uncertainties}, given as
$$
\mathcal{L} = \frac{1}{2HW} \sum_{i=1}^H\sum_{j=1}^W \exp\left(-\log \sigma_{i,j}^2\right) \left\| M_{i,j} - \sigma(f_\theta(X,q))_{i,j} \right\|^2 + \log \sigma_{i,j}^2,
$$
where $\log \sigma^2_{i,j}$ is the predicted log variance at spatial position $(i, j)$, $M$ is the ground-truth mask, $\sigma(\cdot)$ is the sigmoid function applied element-wise to each spatial element, and $f_\theta(X, q)$ is the output of SAM. Unlike LA, this method uses a single backward pass that combines all eight steps.

\paragraph{Hyperparameter settings.} \autoref{tab:hyperparameter_overview} summarises the hyperparameters used during training of the modules in this study. \autoref{tab:augmentation_settings} details the hyperparameters of the data augmentation settings. All training procedures follow the 8-step schedule. Data augmentation is limited to random horizontal flips, mirroring the pre-training setup used in SAM~\citep{ravi2024sam}. We also adopt their bounding box perturbation strategy, adding noise up to 10\% of box dimensions (max 20 pixels). Optimisation uses AdamW~\citep{loshchilov2017decoupled} with constant learning rate scheduling and precision tailored to stability: bfloat16 for the auxiliary variance network and float32 for Laplace-based setups to avoid overflow/underflow during sampling.

\begin{table}[h]
\centering
\footnotesize 
\begin{minipage}[t]{0.5\linewidth}
\raggedright
\begin{tabular}{ll}
\toprule
\textbf{Parameter} & \textbf{Value} \\
\midrule
\textit{Multi-Step Training Procedure} & \\
Num. Steps  &  8   \\
Foreground Probability & 0.5 \\
Step 1: point/bbox probability & 0.5 \\
Sampling Strategy & Uniform \\
Augmentations & Rand. HFlip \\
Max Number of Objects & 32 \\
bbox noise level & 10\% Box dim., \\
& $<=20$ \\ 
\midrule 
\textit{Optimisation} & \\
Optimiser & AdamW \\ 
Learning Rate & 1e-4 \\
Schedule & Constant \\
Weight Decay & 1e-4 (VarNet), 0.01\\
Gradient Clip Norm & 0.1 \\
Batch Size (images) & 1 \\
Steps & 17,500 \\
Warmup steps & 195 \\ 
Cooldown steps & 972 \\ 
\bottomrule
\end{tabular}
\vspace{2mm}
\caption{Overview of training hyperparameters.}
\label{tab:hyperparameter_overview}
\end{minipage}
\hfill
\begin{minipage}[t]{0.45\linewidth}
\raggedright
\begin{tabular}{ll}
\toprule
\multicolumn{2}{l}{\textbf{Training Augmentations}} \\
\midrule
Horizontal Flip (hflip)      & p: 0.5 \\
Resize                       & 1024 (square) \\
\midrule
\multicolumn{2}{l}{\textbf{TTA (Train)}} \\
\midrule
Color Jitter (1)             & b: 0.1, c: 0.03 \\ 
& s: 0.03, h: null \\
Greyscale                    & p: 0.05 \\
Color Jitter (2)             & b: 0.1, c: 0.05 \\ 
& s: 0.05, h: null \\
Resize                       & 1024 (square) \\
\midrule 
\multicolumn{2}{l}{\textbf{TTA (Hue)}} \\
\midrule
Color Jitter                 & b: null, c: null \\ 
& s: null, h: 0.5 \\
Resize                       & 1024 (square) \\
\bottomrule
\end{tabular}
\vspace{2mm}
\caption{Data augmentation sets and corresponding hyperparameters.}
\label{tab:augmentation_settings}
\end{minipage}
\end{table}

\section{Additional Results}
\label{app:more-results}

\begin{figure}[!ht]
    \centering
    \includegraphics[width=0.7\linewidth]{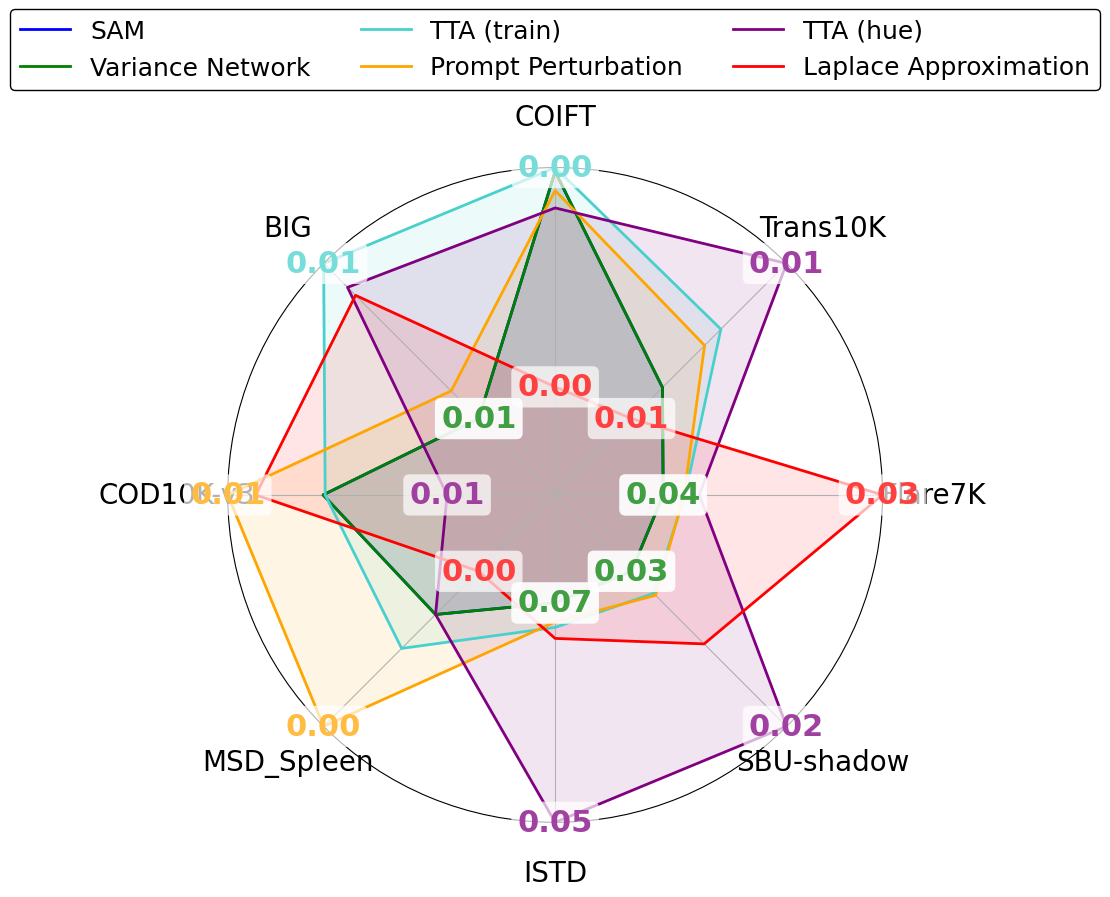}
    \caption{
    Radar plot illustrating Brier scores of the UQ methods, averaged across samples per dataset. Lower scores indicate better probabilistic performance.}
    \label{fig:brier_scores}
\end{figure}

\begin{figure}[ht]
    \centering
    \footnotesize
    \begin{tabular}{@{} m{3cm} m{0.8\linewidth} @{}}
    \textbf{TTA} & \includegraphics[width=\linewidth]{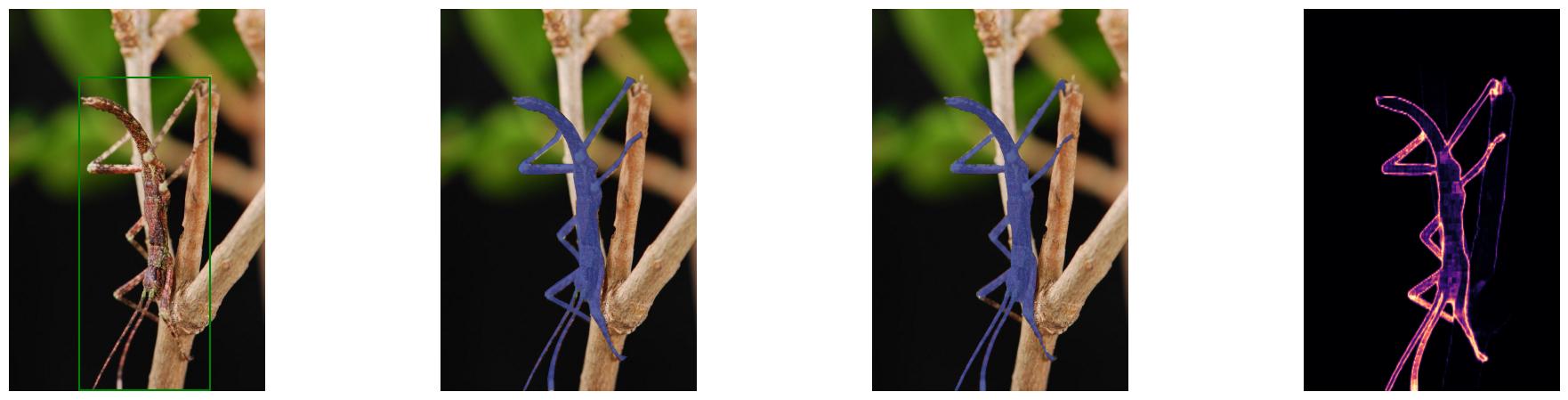} \\[1ex]
    \textbf{Prompt Perturbation} & \includegraphics[width=\linewidth]{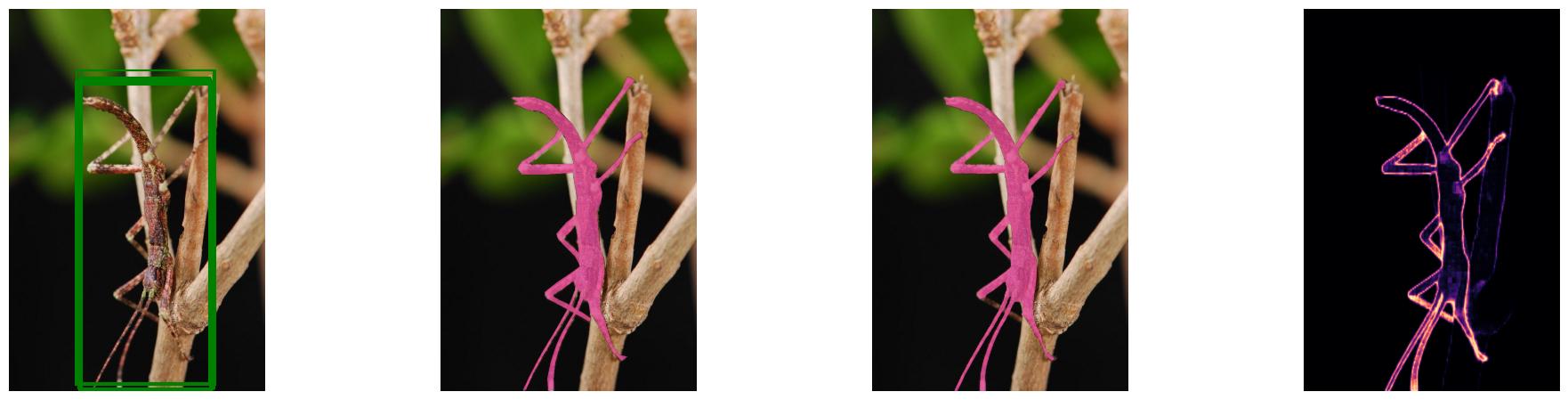} \\[1ex]
    \textbf{Laplace Approx.} & \includegraphics[width=\linewidth]{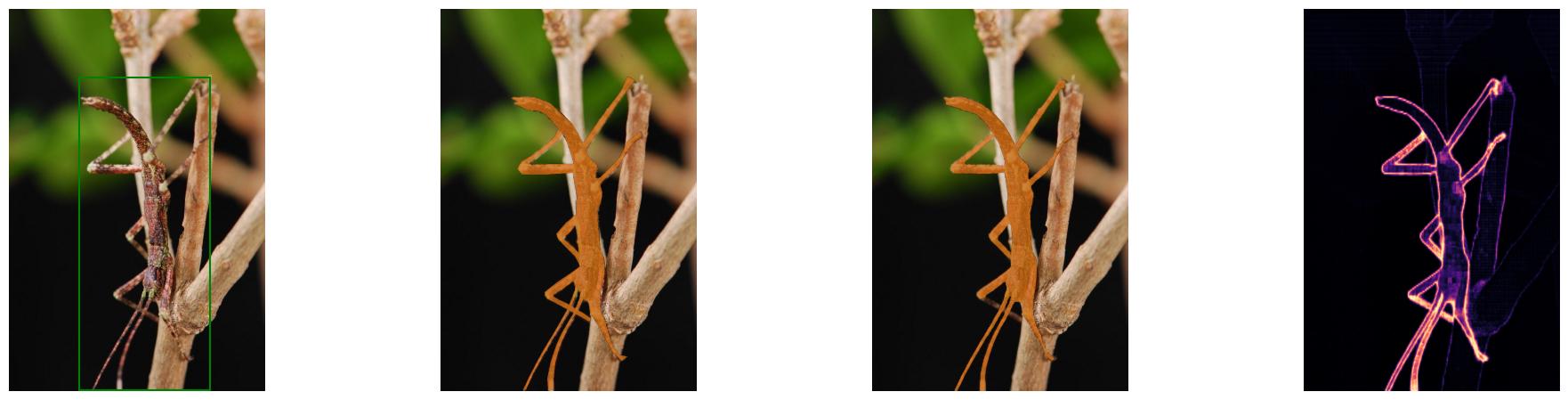} \\[1ex]
    \textbf{Variance Network} & \includegraphics[width=\linewidth]{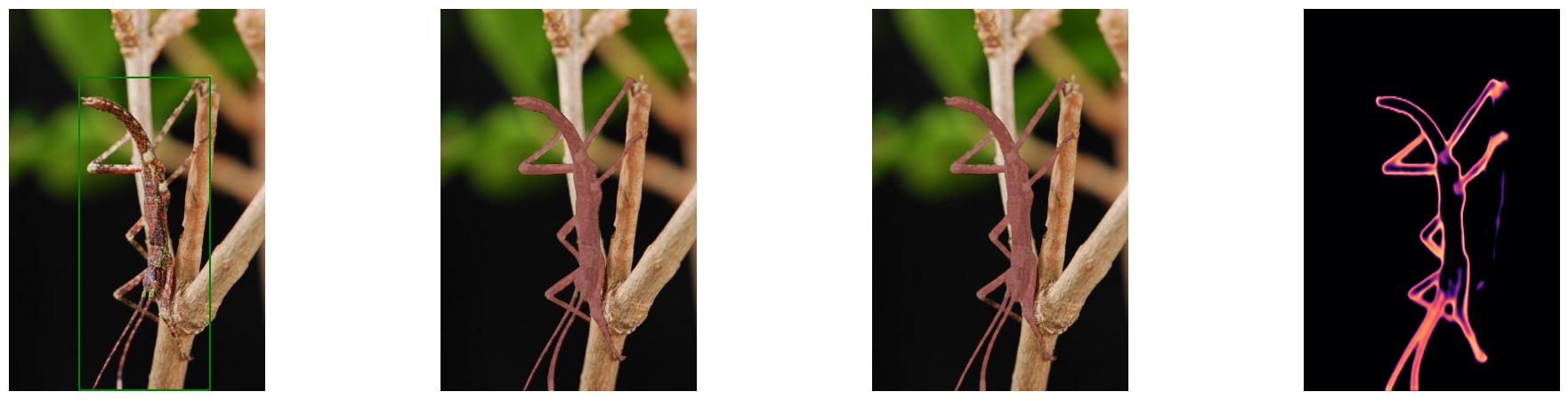}
\end{tabular}

\caption{
A comparative example from the COD camouflage dataset of the four UQ methods. \emph{From left to right in each row}: (1) Input image with bounding box prompt, (2) Ground truth segmentation mask, (3) Predicted segmentation mask, and (4) Uncertainty estimation mask.
}
\label{fig:uncertainty_comparison1}
\end{figure}

\begin{figure}[ht]
    \centering
    \footnotesize
    \begin{tabular}{@{} m{3cm} m{0.8\linewidth} @{}}
    \textbf{TTA} & \includegraphics[width=\linewidth]{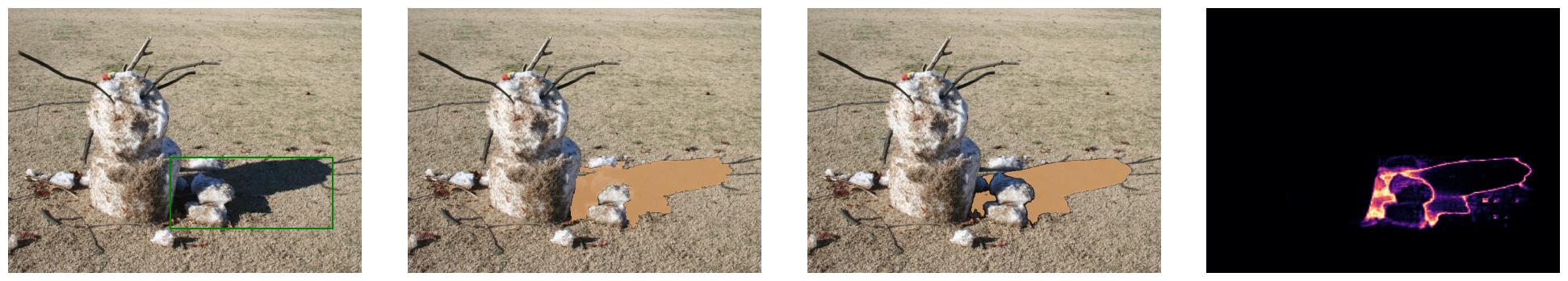} \\[1ex]
    \textbf{Prompt Perturbation} & \includegraphics[width=\linewidth]{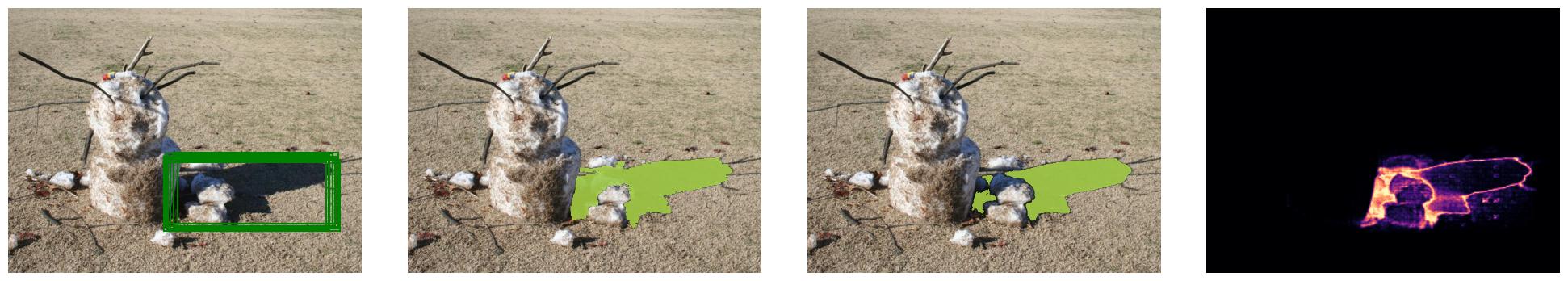} \\[1ex]
    \textbf{Laplace Approx.} & \includegraphics[width=\linewidth]{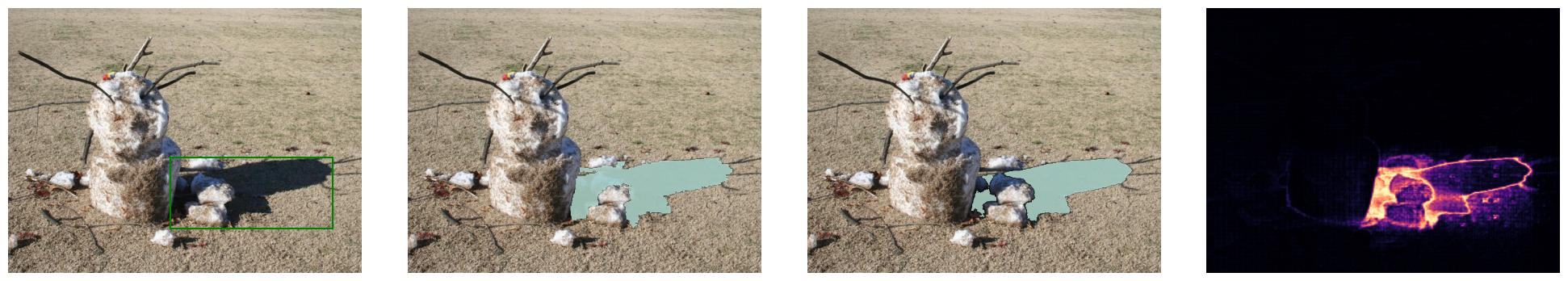} \\[1ex]
    \textbf{Variance Network} & \includegraphics[width=\linewidth]{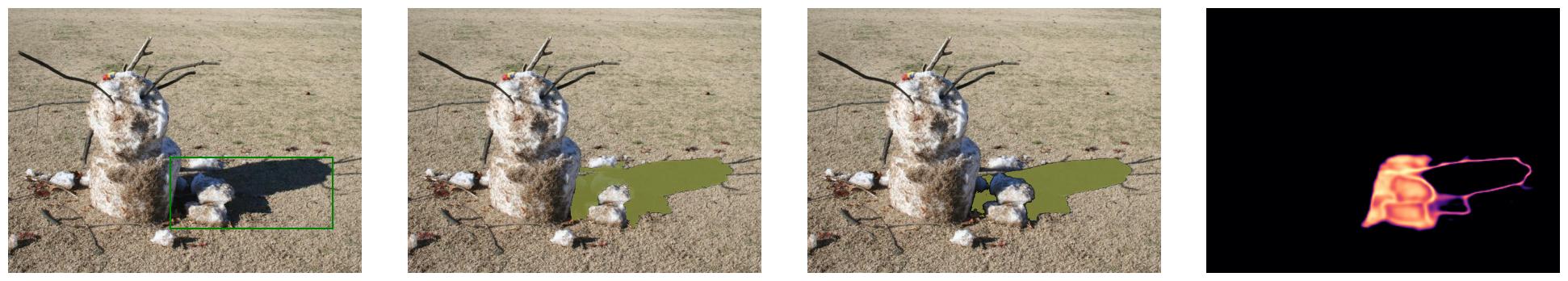}
\end{tabular}

\caption{
A comparative example from the ISTD shadow dataset of the four UQ methods. \emph{From left to right in each row}: (1) Input image with bounding box prompt, (2) Ground truth segmentation mask, (3) Predicted segmentation mask, and (4) Uncertainty estimation mask.
}
\label{fig:uncertainty_comparison2}
\end{figure}

In addition to the correlation analysis in the main text, \autoref{fig:brier_scores} provides a comparison of the \emph{Brier score} \citep{gneiting2007strictly}, a proper scoring rule capturing both calibration and prediction properties. We observe similarly low values across UQ methods, indicating comparable probabilistic accuracy across methods. In addition, \autoref{fig:uncertainty_comparison1} and \autoref{fig:uncertainty_comparison2} provide two qualitative comparisons along the lines of \autoref{fig:istd_ignorance} for the uncertainty maps generated by each of our four considered uncertainty estimation methods.

\end{document}